\documentclass{article}
\usepackage{iclr2027_conference,times}

\usepackage{amsmath,amsfonts,bm}

\def\eqref#1{equation~\ref{#1}}

\def\1{\bm{1}}

\DeclareMathAlphabet{\mathsfit}{\encodingdefault}{\sfdefault}{m}{sl}
\SetMathAlphabet{\mathsfit}{bold}{\encodingdefault}{\sfdefault}{bx}{n}

\usepackage{amsmath,amssymb,amsthm,mathtools}
\usepackage{booktabs}
\usepackage{graphicx}
\usepackage{xcolor}
\usepackage{microtype}
\usepackage{tabularx}
\usepackage{enumitem}
\usepackage{xspace}
\usepackage[hidelinks]{hyperref}
\hypersetup{
  pdftitle={What Memory Composition Does Not Tell Us About Anomaly Detection},
  pdfauthor={Joongwon Chae, Runming Wang, Peiwu Qin}
}
\usepackage{url}

\newcommand{\method}{\textsc{CleanCon}\xspace}

\title{What Memory Composition Does Not Tell Us About Anomaly Detection}
\author{
Joongwon Chae$^{1,3}$ \qquad Runming Wang$^{1}$ \qquad Peiwu Qin$^{2}$\\
\normalfont
$^{1}$Institute of Biopharmaceutical and Health Engineering,\\
Tsinghua Shenzhen International Graduate School, Tsinghua University, Shenzhen, China\\
$^{2}$Guangdong Provincial Laboratory of Traditional Chinese Medicine,\\
Hengqin, Guangdong, China\\
$^{3}$RatelSoft
}

\iclrfinalcopy  

\begin{document}
\maketitle
\lhead{Preprint}

\begin{abstract}
Memory-based anomaly detectors store nominal training patches and score test patches against this memory. A patch selected for coverage therefore becomes a normal reference without a separate check that geometric rarity makes it safe to trust. We probe this coupling with sparse training contamination. Under fixed representations and memory budgets, we compare random, medoid, local, and global coverage selectors. We then use \method, an out-of-bag cross-image support gate that changes candidate-image eligibility while fixing the representation, absolute memory size, builder, and inference rule. Global coverage strongly over-represents sparse contamination. \method reduces final-memory contamination to approximately zero and increases category-macro P-AP in all 12 matched comparisons. Yet along a retention sweep, the lowest-contamination memory does not attain the highest P-AP; performance continues to improve while contamination rises. Memory contamination therefore does not order the resulting memories by P-AP.Code is publicly available at \url{https://github.com/jw-chae/cleancon}.
\end{abstract}

\section{Introduction}
\label{sec:intro}

Memory-based industrial anomaly detection extracts patch features from nominal training images, stores a compact subset in memory, and identifies anomalies through discrepancies between test patches and the stored features~\citep{cohen2020spade,defard2021padim,roth2022patchcore}. In this family, memory is not merely a compressed training set. Selected patches remain direct normal references at inference time.

PatchCore constructs this memory with an approximate-greedy coreset that seeks broad feature-space coverage~\citep{roth2022patchcore}. The sensitivity of coverage objectives to outliers is well known in the literature on $k$-center and robust clustering~\citep{charikar2001facilityoutliers,friedler2010robustkcenter,krishnaswamy2018outliers}. In memory-based anomaly detection, however, this sensitivity has a more direct consequence. A selected center does not disappear into a subsequent optimizer; it remains in the deployed detector as a normal reference.

\begin{figure}[t]
  \centering
  \includegraphics[width=\columnwidth]{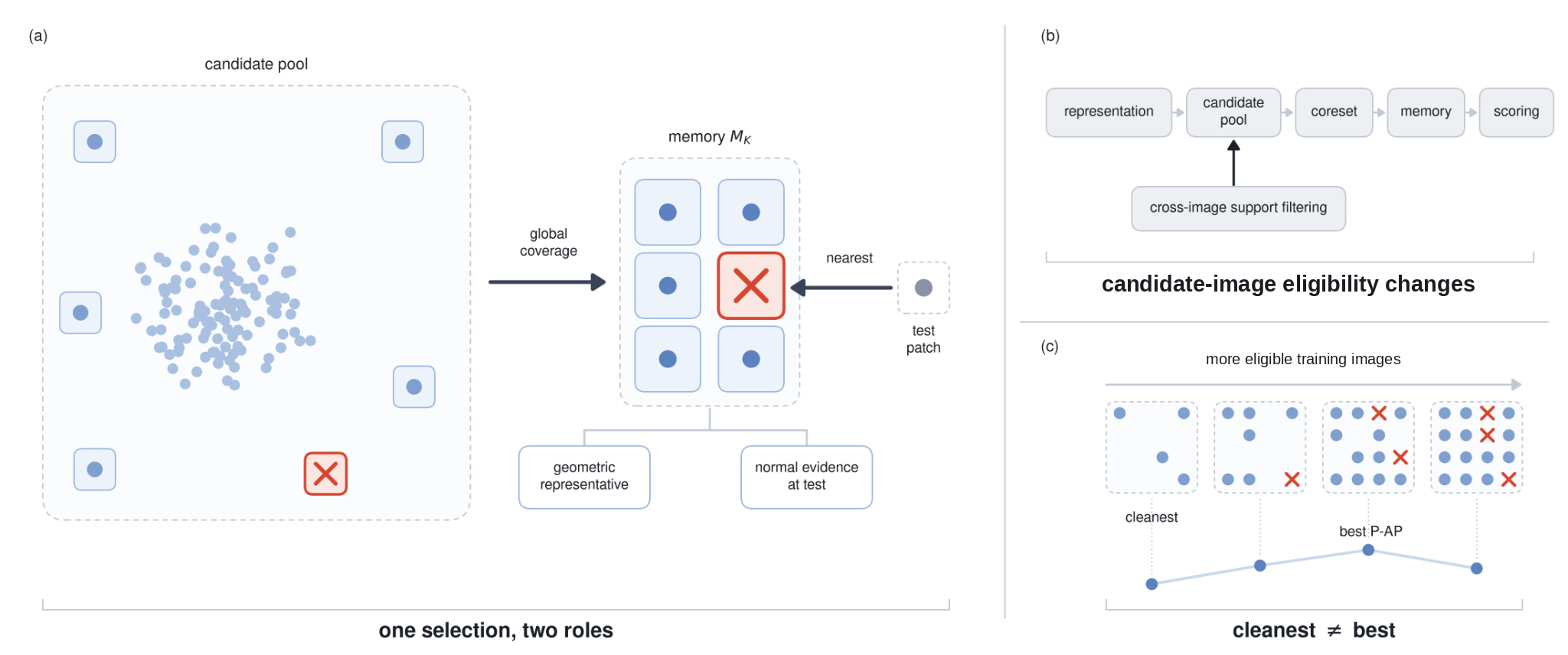}
  \caption{\textbf{One selection, two roles.} (a) Selected patches serve global coverage and test-time normal evidence. (b) Candidate-image eligibility changes while the absolute memory size and detector remain fixed. (c) The lowest-contamination memory is not necessarily the best-performing one.}
  \label{fig:overview}
\end{figure}

Sparse contamination provides a simple probe of this coupling (Fig.~\ref{fig:overview}). If a small number of anomalous images enter the nominal training pool, rare patches within them can be attractive to a global coreset. We ask two questions. How strong is this selection bias in the final memory? And does reducing the resulting contaminated references improve the detector?

We compare random sampling, medoid selection, local approximate-greedy farthest-first selection (Local FF), and global farthest-first selection (Global FF) under identical representations and budgets. \method then ranks each training image by an out-of-bag cross-image projection residual and passes only low-scoring images to the original builder. It changes candidate-image eligibility while fixing the representation, absolute memory size, builder, and inference rule.

Global FF draws substantially more contaminated patches into memory. \method reduces final-memory contamination to approximately zero and increases category-macro P-AP in all 12 matched comparisons. Yet along the retention sweep, the lowest-contamination memory does not attain the highest P-AP, and performance continues to improve as contamination rises. Cleaning helps, but purity alone does not order useful memories.

Our main findings are:
\begin{itemize}[leftmargin=*,nosep]
  \item \textbf{Global coverage strongly over-represents sparse contamination.} Random, medoid, and Local FF controls place the strongest effect under Global FF competition across images.
  \item \textbf{\method reduces contamination and raises matched macro P-AP.} With all downstream components and bank sizes fixed, final-memory contamination approaches zero and category-macro P-AP rises in all 12 comparisons.
  \item \textbf{The cleanest memory is not the best memory.} Along the retention sweep, contamination and P-AP repeatedly rise together.
\end{itemize}

\section{Related Work}
\label{sec:related}

\subsection{Memory-based anomaly detection}
SPADE and PaDiM model nominal patterns with pretrained local features~\citep{cohen2020spade,defard2021padim}, while PatchCore compresses training patches with approximate-greedy coreset selection and computes anomaly scores from nearest-memory distances~\citep{roth2022patchcore}. Later work strengthens representations, adapts features, or introduces projection- and consistency-based scoring~\citep{reiss2021panda,reiss2023meanshift,oquab2023dinov2,chae2026procon}. In these methods, memory composition determines not only inference cost but also which features remain available as normal references.

\subsection{Contaminated nominal training}
Identifying latent outliers and limiting the influence of noisy nominal data are longstanding problems in one-class learning and anomaly detection~\citep{qiu2022loe,yoon2022refine,cordier2022sroc}. For industrial anomaly detection, SoftPatch adjusts memory construction and inference with patch-level outlier scores~\citep{jiang2022softpatch}, while SoftPatch+ combines multiple discriminators~\citep{wang2025softpatchplus}. InReaCh uses cross-image support, FUN-AD iteratively reconstructs a memory bank, and MeDS distills scores from bootstrapped partial memories~\citep{mcintosh2023inreach,im2025funad,safarov2026meds}. SSFilter combines sample-level filtering with uncertainty, while LLB-NAD aggregates scores from multiple models before training a final detector~\citep{liu2026ssfilter,zhang2025llbnad}.

These methods differ in detector, backbone, and contamination construction. Our matched intervention instead fixes the deployed detector while changing candidate-image eligibility, so amplification during memory selection and the detector response after filtering are measured within the same scaffold.

\subsection{Coverage selection and deployed references}
Farthest-first traversal and metric $k$-center seek to cover a space with a limited number of centers~\citep{gonzalez1985clustering}; robust variants restrict the influence of outliers~\citep{charikar2001facilityoutliers,friedler2010robustkcenter,krishnaswamy2018outliers}. More generally, a coreset guarantee depends on the objective it preserves~\citep{harpeled2004coresets,feldman2011framework}, and geometric coverage has been distinguished from downstream performance in active learning and data selection~\citep{sener2018coreset,borsos2021bilevel,mirzasoleiman2020coresets,killamsetty2021gradmatch}.

In memory-based anomaly detection, the selected subset does not disappear into a later optimizer; it remains the inference-time reference set. This makes the path from high coverage value to normal-reference membership, and from that memory to downstream detection, directly observable.

\section{Problem Setup and Experimental Protocol}
\label{sec:setup}

\subsection{Measuring memory contamination}
A frozen encoder $f$ maps training images to patch descriptors, and a memory builder selects $K$ references from the candidate set $\mathcal X$. We denote the selected memory by $\mathcal M_K\subseteq\mathcal X$. For post-hoc analysis, we partition
\begin{equation}
  \mathcal X=\mathcal N\,\dot\cup\,\mathcal Q,
\end{equation}
where $\mathcal Q$ contains descriptors aligned with ground-truth anomalous regions in injected images and $\mathcal N$ contains the remaining candidates. These labels are never supplied to the selector or detector.

We align each ground-truth mask to the $28\times28$ feature grid. A grid cell is labeled contaminated when its corresponding image region contains at least one anomalous pixel; no anomaly-area threshold is applied. One grid cell corresponds to approximately $14\times14$ input pixels for a $392\times392$ DINOv2 input and $8\times8$ pixels for a $224\times224$ WRN50 input. Normal images have all-zero patch labels. The flattened label vector follows feature ordering. For WRN50, the label denotes a position on the final $28\times28$ feature grid rather than the entire convolutional receptive field.

We define candidate-pool prevalence and final-memory occupancy as
\begin{equation}
  p_{\mathrm{pool}}
  =\frac{|\mathcal Q|}{|\mathcal X|},
  \qquad
  p_{\mathrm{mem}}
  =\frac{|\mathcal M_K\cap\mathcal Q|}{K},
\end{equation}
and measure amplification by
\begin{equation}
  A_K
  =\frac{p_{\mathrm{mem}}}{p_{\mathrm{pool}}}
  =\frac{|\mathcal X|\,|\mathcal M_K\cap\mathcal Q|}{K|\mathcal Q|}.
  \label{eq:amplification}
\end{equation}
$A_K=1$ indicates prevalence-proportional selection, while $A_K>1$ indicates that contamination is over-represented in memory. Memory purity is $1-p_{\mathrm{mem}}$.

For DINOv2/ProCon, matched-intervention and retention $p_{\mathrm{mem}}$ pool categories, four depths, and five banks within each fold; selector and radius analyses use one memory per depth (Appendix~\ref{app:amplification}). Means and sample standard deviations are over fold-level ratios, not bank ratios. Candidate counts for $p_{\mathrm{pool}}$ are pooled over categories and depths.

\subsection{No-Overlap contamination protocol}
We use all 15 MVTec AD categories and all 12 VisA categories~\citep{bergmann2019mvtec,zou2022visa}. For each category, a subset of anomalous images is injected into the nominal training set. Nominal $5\%$ contamination denotes the \emph{image-level fraction} of injected anomalous images in the final mixed training set. By contrast, $p_{\mathrm{pool}}$ counts only descriptors aligned with mask-positive regions, and is therefore smaller and category dependent.

Injected anomalous images are removed from evaluation, so memory construction and final evaluation share no anomalous image paths. Each contamination fold changes the injected images and therefore also changes the corresponding No-Overlap evaluation composition. Cross-fold results measure repetition across different contamination memberships and their matched evaluation splits.

The main analyses use nominal $5\%$ image-level contamination and three contamination folds. Injection membership changes across folds while the stochastic conditions of approximate-greedy traversal remain fixed. The default memory contains an absolute $1\%$ of the unfiltered candidate pool. Nested budgets from $0.1\%$ to $10\%$ are used to study memory-size dependence.

\subsection{Detectors and selector controls}
We evaluate two memory-based detector families. The first uses WRN50 representations with PatchCore nearest-memory inference~\citep{roth2022patchcore}; the second uses frozen DINOv2 representations with ProCon projection-consensus inference~\citep{oquab2023dinov2,chae2026procon}. Feature extraction, preprocessing, and test-time scoring follow each detector's default configuration.

Under identical representations and budgets, we compare patch-random, image-balanced random, sampled medoid, Local FF, and Global FF selection. Random controls use 20 repetitions. Full results for deduplication-, density-, and landmark-based selectors appear in the appendix. No selector receives contamination identities or ground-truth patch labels.

We report category-macro image AUROC (I-AUROC), pixel AUROC (P-AUROC), pixel average precision (P-AP), and AUPRO.

For the contaminated-training control, we implement the single-discriminator SoftPatch selection and reliability-weighting rule inside each detector scaffold~\citep{jiang2022softpatch}. In WRN50 rows, both the SoftPatch-style control and \method use PatchCore nearest-memory inference. In DINOv2 rows, both use ProCon projection-consensus inference. The two arms share the representation and inference rule and differ in their memory-construction rule.

In matched \method comparisons, the representation, preprocessing, final memory builder, test-time inference, and \emph{absolute memory size $K$} are fixed. We first set $K$ to $1\%$ of the unfiltered candidate pool and retain the same number of references after filtering. Ground-truth patch labels are not used by OOB scoring or filtering.

\section{Global Coverage Over-Represents Sparse Contamination}
\label{sec:amplification}

\begin{figure*}[t]
  \centering
  \includegraphics[width=\textwidth]{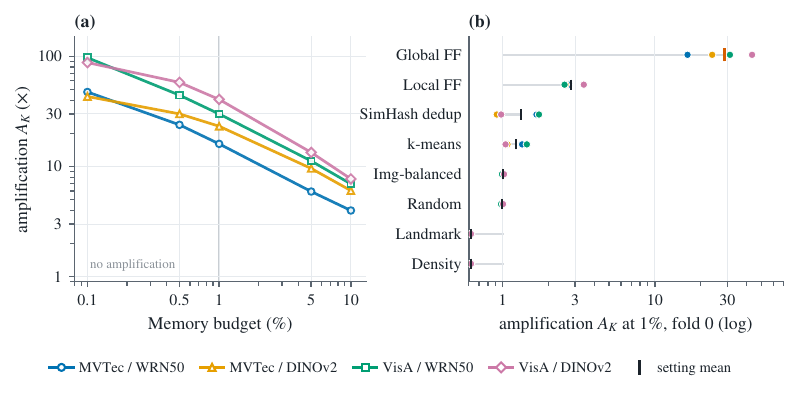}
  \caption{\textbf{Global coverage preferentially retains sparse contamination.} Left: amplification over nested memory budgets from $0.1\%$ to $10\%$. Curves show means over three contamination folds and bands show fold standard deviation. Right: fold-0 selector controls at a $1\%$ memory budget; random controls use 20 repetitions.}
  \label{fig:amplification}
\end{figure*}

\subsection{Amplification depends strongly on the selector}
Figure~\ref{fig:amplification} compares contamination amplification across memory budgets and selectors. At a $1\%$ budget, the three-fold mean amplification of Global FF ranges from $16.04\times$ to $40.61\times$ across the four dataset--representation settings; the largest individual fold reaches $43.57\times$.

In the fold-0 selector control in Fig.~\ref{fig:amplification}b, random remains within $0.99$--$1.02\times$, medoid within $1.05$--$1.44\times$, Local FF within $2.55$--$3.42\times$, and Global FF reaches $16.41$--$43.57\times$. Restricting farthest-first competition to individual images therefore removes most of the amplification.

This pattern agrees with coverage geometry. Selecting another point in a densely represented region removes little uncovered space, whereas a candidate far from the current memory can yield a large coverage gain simply because it is rare. The objective does not distinguish rare normal variation from contamination.

Amplification can also be read as the fraction of contaminated candidates selected. When $K=0.01|\mathcal X|$,
\begin{equation}
  A_K
  =100\,\frac{|\mathcal M_K\cap\mathcal Q|}{|\mathcal Q|}.
\end{equation}
Thus, $A_K=43.57$ means that approximately $43.57\%$ of contaminated candidate descriptors enter the final $1\%$ memory.

Amplification is largest in the sparse-memory regime. The mean curves for all four settings decrease from $0.1\%$ to $10\%$. Because $A_K$ has a $K$-dependent ceiling, we use this curve as a descriptive budget trend. Category-level trajectories appear in Appendix~\ref{app:amplification}.

\subsection{Some contaminated centers are required at the achieved radius}
We next examine whether any selected contaminated centers are required to maintain the achieved covering radius. Let $S_K$ denote the returned centers and
\begin{equation}
  r_K=\max_{x\in\mathcal X}d(x,S_K).
\end{equation}
Even when every normal candidate is available as a center, define the minimum number of contaminated centers required to cover $\mathcal Q$ at scale $r$ as
\begin{equation}
  \mathcal C_r(\mathcal Q\mid\mathcal N)
  =\min_{C\subseteq\mathcal Q}
  \left\{|C|:\mathcal Q\subseteq B(\mathcal N,r)\cup B(C,r)\right\}.
\end{equation}
Because $S_K\cap\mathcal Q$ is feasible at scale $r_K$,
\begin{equation}
  |S_K\cap\mathcal Q|
  \geq
  \mathcal C_{r_K}(\mathcal Q\mid\mathcal N).
  \label{eq:radius_bound}
\end{equation}

We compute a certified packing lower bound $L_{r_K}$ for $\mathcal C_{r_K}$. It is positive in all four pooled fold-0 settings and directly certifies $0.58$--$6.32\%$ of selected contaminated centers. Thus, at least some contaminated centers are required to maintain the achieved radius, while the same certificate does not establish necessity for the remainder. The full construction appears in Appendix~\ref{app:amplification}.

The selector comparison and radius analysis both show global coverage drawing contamination into memory. We next reduce these references and measure the detector response.

\section{CleanCon: A Controlled Intervention That Reduces Contaminated References}
\label{sec:method}

If global coverage over-selects contaminated patches, the next question is whether reducing these references improves detection. \method leaves the builder unchanged and changes which training images may contribute candidates.

Each training image is scored by how well its features can be explained by features from other training images. Only low-scoring images are passed to the original memory builder, while the representation, final memory size $K$, builder, and test-time inference remain unchanged.

\subsection{Out-of-bag support score}
For a category with $n$ training images, we construct $B=20$ support banks. Bank $b$ contains $\lceil0.20n\rceil$ images with index set $I_b$. The same $I_b$ is shared across feature depths, while descriptor banks are built separately for each layer:
\begin{equation}
  \mathcal M_b^\ell
  =\{z_{j,p}^{\ell}:j\in I_b,\ p\in\mathcal P\}.
\end{equation}
The support banks are built from the same potentially contaminated training pool rather than from a separate clean set.

Image $i$ is evaluated only against banks that exclude it:
\begin{equation}
  \mathcal B_i=\{b\mid i\notin I_b\}.
\end{equation}
This restriction prevents the target image from using its own descriptors as support~\citep{breiman2001randomforests}. For descriptor $z_{i,p}^{\ell}$, we retrieve the $k=5$ nearest descriptors $m_j$ in each $b\in\mathcal B_i$ and compute
\begin{equation}
  d_j=\|z_{i,p}^{\ell}-m_j\|_2.
\end{equation}
The neighbor weights are
\begin{equation}
  w_{i,p,j}^{\ell,b}
  =\frac{\exp(-d_j^2/\tau)}{\sum_{t=1}^{k}\exp(-d_t^2/\tau)},
\end{equation}
and the soft projection and residual are
\begin{equation}
  \widehat z_{i,p}^{\ell,b}
  =\sum_{j=1}^{k}w_{i,p,j}^{\ell,b}m_j,
  \qquad
  r_{i,p}^{\ell,b}
  =\|z_{i,p}^{\ell}-\widehat z_{i,p}^{\ell,b}\|_2.
\end{equation}

The temperature $\tau$ is calibrated from the distance scale of a deterministic descriptor subsample. Downstream metrics are not used for calibration. Additional implementation settings appear in Appendix~\ref{app:method}.

We first take the median residual over OOB banks within each layer and then average across depths:
\begin{equation}
  \mu_{i,p}
  =\frac{1}{|\mathcal L|}
  \sum_{\ell\in\mathcal L}
  \operatorname*{median}_{b\in\mathcal B_i}
  r_{i,p}^{\ell,b}.
\end{equation}

\subsection{Image-level filtering}
Patch residuals are aggregated into an image score by averaging the largest $0.5\%$:
\begin{equation}
  a_i
  =\operatorname{TopMean}_{0.5\%}
  \left(\{\mu_{i,p}\}_{p\in\mathcal P}\right).
\end{equation}
On a $28\times28$ grid, approximately four locations contribute. Within each category, only images at or below the median score are retained:
\begin{equation}
  \mathcal T
  =\left\{i\mid a_i\leq\operatorname*{median}_{j}a_j\right\}.
\end{equation}
All descriptors from images in $\mathcal T$ enter subsequent memory construction. Filtering operates at the image level rather than deleting individual patches. The same median rule is used in every setting independently of the nominal $5\%$ contamination ratio.

\subsection{Matched memory construction}
The final memory size remains fixed after filtering. We first set the absolute target $K$ to $1\%$ of the unfiltered pool and select the same number of references with \method:
\begin{equation}
  |\mathcal M_K^{\mathrm{baseline}}|
  =|\mathcal M_K^{\mathrm{CleanCon}}|
  =K.
\end{equation}
The representation, $K$, memory builder, and test-time inference are identical between baseline and \method. WRN50/PatchCore retains nearest-memory inference, while DINOv2/ProCon retains projection-consensus inference~\citep{roth2022patchcore,chae2026procon}. Ground-truth patch labels are not supplied to OOB scoring or filtering and are used only for composition analysis after memory construction.

Complete implementation settings and pseudocode appear in Appendix~\ref{app:method}.

\section{The Cleanest Memory Is Not Always the Best Memory}
\label{sec:utility}

\subsection{CleanCon reduces contamination and increases macro P-AP}
The unfiltered baseline and \method use the same representation, memory size, builder, and inference rule. We compare all 15 MVTec AD categories and 12 VisA categories, WRN50/PatchCore and DINOv2/ProCon, and three contamination folds. For each fold, we compute category-macro P-AP and pooled $p_{\mathrm{mem}}$, then report the mean and sample standard deviation over folds.

\begin{table*}[t]
\centering
\caption{\textbf{Three-fold matched unfiltered-to-\method comparison.}}
\label{tab:matched}
\scriptsize
\setlength{\tabcolsep}{3.5pt}
\begin{tabular}{lccccc}
\toprule
Setting & Vanilla $p_{\mathrm{mem}}$ & \method $p_{\mathrm{mem}}$ & Vanilla P-AP & \method P-AP & $\Delta$P-AP \\
\midrule
MVTec--WRN50  & $6.481\!\pm\!0.370\%$ & $0.000\%$ & $57.704\!\pm\!2.033$ & $61.141\!\pm\!0.433$ & $+3.438\!\pm\!1.625$ \\
MVTec--DINOv2 & $9.462\!\pm\!0.611\%$ & $0.000\%$ & $69.342\!\pm\!0.995$ & $72.573\!\pm\!0.373$ & $+3.231\!\pm\!0.645$ \\
VisA--WRN50   & $3.678\!\pm\!0.500\%$ & $0.003\!\pm\!0.003\%$ & $39.345\!\pm\!0.586$ & $39.725\!\pm\!0.309$ & $+0.380\!\pm\!0.429$ \\
VisA--DINOv2  & $4.748\!\pm\!0.814\%$ & $0.002\!\pm\!0.000\%$ & $47.271\!\pm\!0.901$ & $51.590\!\pm\!1.821$ & $+4.319\!\pm\!1.139$ \\
\bottomrule
\end{tabular}
\end{table*}

\method increases P-AP in all 12 dataset--representation--fold macro comparisons while reducing final-memory contamination to approximately zero in every setting. The three-fold mean gains are $+3.438$, $+3.231$, $+0.380$, and $+4.319$ points for MVTec--WRN50, MVTec--DINOv2, VisA--WRN50, and VisA--DINOv2, respectively.

The distribution of macro gains differs across settings. At category--fold level, \method is higher in 89 of 162 comparisons and in 13 of 36 comparisons on VisA--WRN50. VisA--WRN50 also has the lowest absolute P-AP and the smallest macro improvement, and is the only setting in Sec.~\ref{sec:baseline-comparison} where \method trails the SoftPatch-style control.

Here, $p_{\mathrm{mem}}\approx0$ describes the memory after both the OOB gate and the final builder. A contaminated image may pass the gate without contributing a labeled contaminated descriptor to the final memory.

\subsection{Contamination and P-AP can rise together as retention increases}
\method raises category-macro P-AP in all 12 matched runs. We next retain the lowest-scoring $R\in\{50,60,70,80,90\}\%$ images as candidates (R50--R90), keeping the builder and final memory size fixed. All four settings use three contamination folds.

\begin{table*}[t]
\centering
\caption{\textbf{Retention sweep: P-AP / pooled memory contamination.} Values are three-fold means.}
\label{tab:retention}
\scriptsize
\setlength{\tabcolsep}{2.5pt}
\begin{tabular}{lccccc}
\toprule
Setting & R50 & R60 & R70 & R80 & R90 \\
\midrule
MVTec--WRN50  & $61.141/0\%$ & $61.717/0\%$ & $61.720/0.0067\%$ & $62.070/0.0089\%$ & $62.084/0.0234\%$ \\
MVTec--DINOv2 & $72.573/0\%$ & $72.743/0\%$ & $73.027/0.0034\%$ & $73.094/0.0100\%$ & $73.144/0.0371\%$ \\
VisA--WRN50   & $39.725/0.0033\%$ & $40.345/0.0112\%$ & $40.453/0.0210\%$ & $40.600/0.0462\%$ & $40.475/0.1460\%$ \\
VisA--DINOv2  & $51.590/0.0018\%$ & $53.509/0.0066\%$ & $53.762/0.0146\%$ & $53.883/0.0270\%$ & $53.155/0.0947\%$ \\
\bottomrule
\end{tabular}
\end{table*}

R80 exceeds R50 in all 12 dataset--representation--fold comparisons. The setting-level three-fold mean differences are $+0.929$, $+0.521$, $+0.875$, and $+2.293$ points. In 11 of the 12 comparisons, R80 also contains more contamination. In the remaining comparison, both operating points have zero contamination and R80 still has higher P-AP.

The full sweep shows the same tension. On MVTec, mean P-AP rises from R50 to R60 while pooled contamination remains zero, then continues to rise after contamination reappears. On VisA, P-AP and contamination rise together through R80 before P-AP falls at R90. More candidates are not always better, but the cleanest tested memory is not best either.

Category-level patterns also vary by setting. R80 exceeds R50 in 11 of 15 MVTec--DINOv2 categories (median $+0.407$), 11 of 12 VisA--DINOv2 categories (median $+2.060$), 9 of 15 MVTec--WRN50 categories, and 7 of 12 VisA--WRN50 categories (median $+0.145$). Removing the largest positive outlier leaves a mean gain of $+0.292$ on MVTec--WRN50. On VisA--WRN50, removing pcb4 changes the mean to $-0.156$.

\begin{figure}[t]
  \centering
  \includegraphics[width=\columnwidth]{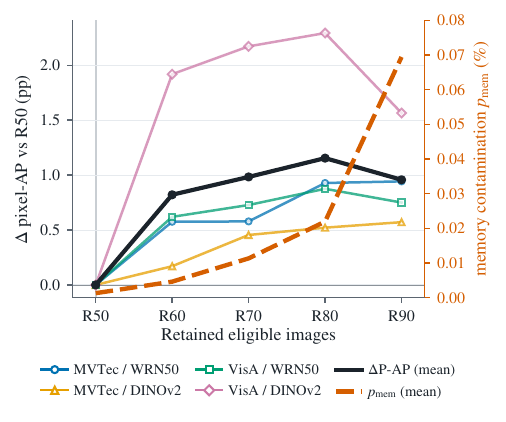}
  \caption{\textbf{The cleanest memory does not attain the highest P-AP.} The left axis shows the change in category-macro P-AP relative to R50 for the four dataset--representation settings and their mean. The right axis shows the setting mean of pooled $p_{\mathrm{mem}}$. R80 exceeds R50 in all 12 comparisons, and contamination is also higher in 11 of them.}
  \label{fig:retention}
\end{figure}

\subsection{Increasing the number of normal sources does not reproduce the gain}
Increasing image retention admits both normal and contaminated candidates. To test whether more normal sources alone explain the gain, we define the fraction of normal training images that contribute at least one descriptor to the final memory:
\begin{equation}
  R_{\mathrm{src}}
  =\frac{|\{i\in\mathcal I_N:\exists m\in\mathcal M_K\text{ sourced from }i\}|}{|\mathcal I_N|}.
\end{equation}

The fixed-budget normal-source expansion control (Appendix~\ref{app:source-expansion}) increases $R_{\mathrm{src}}$ by 20.88 percentage points and changes P-AP by $-0.137\pm0.024$ points. On an auxiliary clean six-category subset, the same 50\% image-retention rule reduces P-AP by 0.939 points.

These results show that the retention gain is not explained by the number of normal sources alone. They leave open the possibility that the identity of the retained normal states and their local feature-space configuration are also relevant.

\subsection{Comparison with noisy-training baselines}
\label{sec:baseline-comparison}
Table~\ref{tab:app-softpatch} compares \method with a single-discriminator SoftPatch selection and reliability-weighting rule under matched detector scaffolds~\citep{jiang2022softpatch}. WRN50 rows share PatchCore inference and DINOv2 rows share ProCon inference. \method has higher macro P-AP in five of six settings; on VisA--WRN50, the SP-style control and \method score 41.495 and 40.061 with pooled memory contamination of approximately $0.108\%$ and $0.006\%$, respectively.

Table~\ref{tab:nooverlap-main} places \method alongside the $5\%$ No-Overlap results reported for InReaCh, SoftPatch, and FUN-AD~\citep{im2025funad}. Their $5\%$ anomaly-to-normal ratio corresponds to approximately $4.762\%$ of the final mixed training set; \method uses approximately $5\%$ mixed-training contamination, fold 0, and seed 0.

\begin{table*}[t]
\centering
\scriptsize
\caption{\textbf{Performance comparison on contaminated MVTec AD and VisA under No-Overlap.}}
\label{tab:nooverlap-main}
\setlength{\tabcolsep}{4.2pt}
\begin{tabular}{lllrrrr}
\toprule
Dataset & Method & Mixed-train contamination & I-AUROC & P-AUROC & P-AP & AUPRO \\
\midrule
MVTec & InReaCh & $\approx4.762\%$ & 92.32 & 97.01 & -- & -- \\
       & SoftPatch & $\approx4.762\%$ & 98.40 & 98.17 & -- & -- \\
       & FUN-AD & $\approx4.762\%$ & 98.45 & 98.43 & -- & -- \\
       & \method--WRN50 & $\approx5\%$ & 97.539 & -- & 61.016 & 93.539 \\
       & \method--DINOv2 & $\approx5\%$ & 99.147 & -- & 72.255 & 95.770 \\
\midrule
VisA   & InReaCh & $\approx4.762\%$ & 83.66 & 97.67 & -- & -- \\
       & SoftPatch & $\approx4.762\%$ & 89.82 & 98.57 & -- & -- \\
       & FUN-AD & $\approx4.762\%$ & 93.65 & 98.92 & -- & -- \\
       & \method--WRN50 & $\approx5\%$ & 92.596 & -- & 40.061 & 93.693 \\
       & \method--DINOv2 & $\approx5\%$ & 97.712 & -- & 49.522 & 96.861 \\
\bottomrule
\end{tabular}
\end{table*}

\section{What Is Stored and What Is Useful Are Different}
\label{sec:discussion}

Global coverage draws sparse contamination into memory; \method reduces it and raises matched category-macro P-AP. The retention sweep nevertheless places the cleanest memory and highest P-AP at different operating points. $A_K$ and $p_{\mathrm{mem}}$ describe what entered memory; the detector scores how the retained references relate to test patches. Memories with the same contamination rate can retain different normal states, densities, and local geometries.

Increasing image retention admits normal and contaminated candidates together, so the sweep does not isolate purity alone. It asks instead whether cleanliness selects the best operating point. In all four settings, memories built from broader candidate pools outperform the cleanest tested memory over part of the sweep, while more normal source images alone do not reproduce the gain. This suggests that source identity and local configuration may also matter.

Memory construction is therefore more than compression. Coverage selects feature-space representatives, filtering controls which training evidence remains, and inference determines how well that memory explains test samples. These stages do not induce the same ordering over memories.

\section{Limitations and Conclusion}
\label{sec:conclusion}

\method changes candidate-image eligibility; selecting rare valid patches without restoring anomalous anchors remains open. The achieved-radius certificate directly covers $0.58$--$6.32\%$ of selected contamination, and retention is evaluated at five operating points with fixed hyperparameters.

Global coverage draws sparse contamination into memory. \method drives final-memory contamination to approximately zero and raises category-macro P-AP in all 12 matched comparisons. Yet P-AP can keep rising as contamination reappears, while increasing normal-source count does not reproduce the gain. Useful memories therefore have structure beyond purity alone.

\newpage

\bibliography{references}
\bibliographystyle{iclr2027_conference}

\appendix
\setcounter{equation}{0}
\renewcommand{\theequation}{A\arabic{equation}}
\setcounter{table}{0}
\renewcommand{\thetable}{A\arabic{table}}
\setcounter{figure}{0}
\renewcommand{\thefigure}{A\arabic{figure}}

\renewcommand{\theHequation}{appendix.\arabic{equation}}
\renewcommand{\theHtable}{appendix.\arabic{table}}
\renewcommand{\theHfigure}{appendix.\arabic{figure}}

\section{Experimental Protocol}
\label{app:experimental}

\subsection{Contamination protocols and No-Overlap evaluation}
Contaminated training sets are constructed by injecting anomalous images into clean training images. Let $n_c$ and $n_a$ denote the numbers of clean and injected anomalous training images. We use two injection protocols.

\paragraph{Exact-Final.}
The fraction of anomalous images in the final mixed training set is
\begin{equation}
  \rho_{\mathrm{final}}
  =\frac{n_a}{n_c+n_a}.
\end{equation}
Under Exact-Final $5\%$, $n_a$ is chosen so that anomalous images constitute approximately $5\%$ of the mixed training set after integer rounding. The amplification analysis, selector comparison, matched intervention, retention sweep, and SoftPatch-style comparison use this protocol.

\paragraph{Addition.}
The injection amount is defined relative to the clean training set:
\begin{equation}
  \rho_{\mathrm{add}}
  =\frac{n_a}{n_c}.
\end{equation}
The corresponding prevalence in the final mixed set is
\begin{equation}
  \rho_{\mathrm{final}}
  =\frac{\rho_{\mathrm{add}}}{1+\rho_{\mathrm{add}}}.
\end{equation}
Addition-$0.10$ therefore corresponds to approximately $9.09\%$ final contamination before integer rounding. The normal-source expansion control and the fold-stability analysis use Addition-$0.10$.

Both protocols use No-Overlap evaluation. Anomalous images injected into training are removed from the downstream evaluation split. Each contamination fold injects a different set of anomalous images and uses the corresponding evaluation split. We report the mean and sample standard deviation across three folds; selector stochastic conditions are fixed unless stated otherwise.

Under Exact-Final $5\%$, each MVTec fold contains 3,629 clean training images and 193 injected anomalous images. The initial anomalous evaluation pool contains 1,065 images, and category-level realized contamination ranges from $4.7619\%$ to $5.1948\%$. Each VisA fold contains 8,659 clean images and 457 injected anomalous images. The initial anomalous evaluation pool contains 743 images, and category-level realized contamination ranges from $4.9578\%$ to $5.0788\%$.

\subsection{Patch labels and aggregation}
Ground-truth masks are aligned to the $28\times28$ feature grid. A descriptor is labeled contaminated if its corresponding image region contains at least one anomalous pixel; no anomaly-area threshold is used. One grid cell corresponds to approximately $14\times14$ input pixels for DINOv2 at $392\times392$ and $8\times8$ pixels for WRN50 at $224\times224$. For WRN50, the label denotes the final feature-grid position.

Normal images have all-zero patch labels, and the label grid is flattened in feature order. Candidate counts are pooled over categories and depths. Writing $n^{\mathrm{anom}}_{c,\ell,b}$ and $n^{\mathrm{all}}_{c,\ell,b}$ for stored descriptor counts, DINOv2/ProCon matched-intervention and retention results compute $p_{\mathrm{mem}}=(\sum_{c,\ell,b}n^{\mathrm{anom}}_{c,\ell,b})/(\sum_{c,\ell,b}n^{\mathrm{all}}_{c,\ell,b})$ over all categories, four depths, and five banks within each fold. We report the mean and sample standard deviation of the three fold-level ratios, not an average of bank ratios. Selector and radius analyses use one memory per depth (Appendix~\ref{app:amplification}).

Detection metrics are computed per category and then macro-averaged. Three-fold summaries first compute each fold's category-macro metric and then report the mean and sample standard deviation over folds.

\subsection{Detectors, memory budgets, and evaluation metrics}
\begin{center}
\begin{tabular}{lll}
\toprule
Setting & Representation & Memory / inference \\
\midrule
WRN50 & frozen WRN50 features & PatchCore-style / nearest reference \\
DINOv2 & frozen DINOv2 features & ProCon projection consensus \\
\bottomrule
\end{tabular}
\end{center}

The default final memory size is an absolute $1\%$ of the unfiltered candidate pool. The same $K$ is retained after \method filtering. Budget-dependent amplification is evaluated on nested prefixes of $0.1\%$, $0.5\%$, $1\%$, $5\%$, and $10\%$.

We report I-AUROC, P-AUROC, P-AP, and AUPRO. The main text uses P-AP as the primary localization metric.

\subsection{Selector controls}
We compare the following selectors under identical representations and memory budgets:
\begin{enumerate}[leftmargin=*]
  \item patch-level random sampling,
  \item image-balanced random sampling,
  \item SimHash deduplication,
  \item density top-$K$,
  \item sampled $k$-means medoid,
  \item Voronoi landmark mass,
  \item local farthest-first selection (Local FF), and
  \item global farthest-first selection (Global FF).
\end{enumerate}
Patch random samples uniformly from the pooled candidate set. Image-balanced random assigns samples to source images in round-robin order. Both random controls use 20 repetitions. Local FF divides memory capacity across images and then performs farthest-first selection within each image. All selectors use the same candidate representation and absolute memory size.

\section{CleanCon Implementation}
\label{app:method}

\subsection{Out-of-bag banks and patch scores}
For a category with $n$ training images, we independently sample $B=20$ image-membership sets $I_b$, each containing $\lceil0.20n\rceil$ images. Membership is shared across feature depths, while descriptor banks are layer specific:
\begin{equation}
  \mathcal M_b^\ell
  =\{z_{j,p}^{\ell}:j\in I_b,\ p\in\mathcal P\}.
\end{equation}
Target image $i$ uses only banks that exclude it:
\begin{equation}
  \mathcal B_i=\{b:i\notin I_b\}.
\end{equation}
For each layer and patch, we retrieve the $k=5$ nearest descriptors $m_j$ and compute
\begin{equation}
  d_j=\|z_{i,p}^{\ell}-m_j\|_2.
\end{equation}
The soft projection weights and projected descriptor are
\begin{equation}
  w_{i,p,j}^{\ell,b}
  =\frac{\exp(-d_j^2/\tau)}{\sum_{t=1}^{k}\exp(-d_t^2/\tau)},
  \qquad
  \widehat z_{i,p}^{\ell,b}
  =\sum_{j=1}^{k}w_{i,p,j}^{\ell,b}m_j,
\end{equation}
and the residual is
\begin{equation}
  r_{i,p}^{\ell,b}
  =\|z_{i,p}^{\ell}-\widehat z_{i,p}^{\ell,b}\|_2.
\end{equation}
The temperature $\tau$ is calibrated from the distance scale of a deterministic descriptor subsample. Residuals are aggregated by taking the OOB-bank median within each layer and then averaging across depths:
\begin{equation}
  \mu_{i,p}
  =\frac{1}{|\mathcal L|}
  \sum_{\ell\in\mathcal L}
  \operatorname*{median}_{b\in\mathcal B_i}
  r_{i,p}^{\ell,b}.
\end{equation}

\subsection{Image ranking and retention}
The image score is the mean of the highest $0.5\%$ patch residuals:
\begin{equation}
  a_i
  =\operatorname{TopMean}_{0.5\%}
  \left(\{\mu_{i,p}\}_{p\in\mathcal P}\right).
\end{equation}
On the $28\times28$ grid, approximately four locations contribute. Default \method retains images at or below the category median:
\begin{equation}
  \mathcal T
  =\{i:a_i\leq\operatorname*{median}_{j}a_j\}.
\end{equation}
All descriptors from images in $\mathcal T$ are passed to the original memory builder. The retention sweep uses the same ranking and retains $R\in\{50,60,70,80,90\}\%$ of images.

\subsection{Fixed-budget construction}
Vanilla and \method share the representation, preprocessing, Global FF builder, inference rule, and absolute memory size:
\begin{equation}
  |\mathcal M_K^{\mathrm{Vanilla}}|
  =|\mathcal M_K^{\mathrm{CleanCon}}|
  =K.
\end{equation}
The DINOv2/ProCon path uses four feature depths, five equal-size seed-perturbed memory banks per depth, a 192-dimensional selection projection, $k=5$ soft local projection, fp16 memory storage, and Gaussian map smoothing with $\sigma=4$. Vanilla and \method use the same target count for each layer--bank pair. Matched-intervention and retention $p_{\mathrm{mem}}$ pools all pairs; selector and radius counts use one memory per depth (Appendix~\ref{app:amplification}). The WRN50/PatchCore path uses a 128-dimensional selection projection and PatchCore inference.

The approximate-greedy builder samples 10 auxiliary points with NumPy RNG seed 0. Candidate distance is initialized by the mean Euclidean distance to the auxiliary points and subsequently updated by the distance to the most recently selected center.

\subsection{Pseudocode}
\begin{enumerate}[leftmargin=*]
  \item Extract multi-depth patch descriptors $z_{i,p}^{\ell}$ from all training images.
  \item Sample $B=20$ image memberships $I_1,\ldots,I_B$.
  \item For each image $i$, compute soft-projection residuals using banks with $i\notin I_b$.
  \item Aggregate by OOB-bank median and depth mean to obtain $\mu_{i,p}$.
  \item Average the highest $0.5\%$ patch residuals to obtain $a_i$.
  \item Retain the lowest-scoring $R\%$ images within each category.
  \item Pass all descriptors from retained images to the original memory builder.
  \item Build each final memory bank at the fixed target size and run the original inference.
\end{enumerate}

\section{How Global Coverage Selects Contamination}
\label{app:amplification}

\subsection{Budget-dependent amplification}
For category $c$ and budget $b$, let $n_{c,b}$, $q_c$, $K_{c,b}$, and $m_{c,b}$ denote the candidate count, contaminated candidate count, memory size, and selected contaminated count. Micro-pooled amplification is
\begin{equation}
  A_b^{\mathrm{micro}}
  =\frac{\sum_c m_{c,b}/\sum_c K_{c,b}}{\sum_c q_c/\sum_c n_{c,b}}.
\end{equation}
DINOv2 selector and radius analyses pool categories and four depths for one seed-0 memory per depth; matched-intervention and retention $p_{\mathrm{mem}}$ pool all five banks (Sec.~\ref{sec:setup}). WRN50 uses a merged patch representation.

\begin{table*}[t]
\centering
\scriptsize
\caption{Micro-pooled amplification over nested memory budgets. Values are mean $\pm$ sample standard deviation over contamination folds 0--2.}
\label{tab:app-budget}
\begin{tabular}{lccccc}
\toprule
Setting & $0.1\%$ & $0.5\%$ & $1\%$ & $5\%$ & $10\%$ \\
\midrule
MVTec--WRN50 & $47.47\pm2.82$ & $23.88\pm1.81$ & $16.04\pm0.77$ & $5.92\pm0.21$ & $3.99\pm0.14$ \\
VisA--WRN50 & $97.15\pm5.90$ & $44.39\pm2.82$ & $29.97\pm1.25$ & $11.22\pm0.31$ & $6.94\pm0.18$ \\
MVTec--DINOv2 & $43.17\pm2.97$ & $30.01\pm2.31$ & $23.19\pm1.67$ & $9.59\pm0.60$ & $6.04\pm0.32$ \\
VisA--DINOv2 & $87.73\pm5.89$ & $57.99\pm4.02$ & $40.61\pm2.65$ & $13.41\pm0.55$ & $7.70\pm0.26$ \\
\bottomrule
\end{tabular}
\end{table*}

Among 162 category-level trajectories, 156 are non-increasing on the coarse budget grid. All six increases occur between $0.1\%$ and $0.5\%$.

\begin{table*}[t]
\centering
\scriptsize
\caption{Fold-0 Global FF counts at a $1\%$ per-memory budget. DINOv2 pools categories and four depths for one seed-0 memory per depth.}
\label{tab:app-counts}
\begin{tabular}{lrrrrrr}
\toprule
Setting & $n$ & $q$ & $K$ & $m_K$ & $A_K$ & Selected $Q$ \\
\midrule
MVTec--WRN50 & 2,996,448 & 12,586 & 29,958 & 2,065 & 16.41 & 16.41\% \\
VisA--WRN50 & 7,146,944 & 7,210 & 71,464 & 2,247 & 31.17 & 31.17\% \\
MVTec--DINOv2 & 11,985,792 & 51,036 & 119,832 & 12,160 & 23.83 & 23.83\% \\
VisA--DINOv2 & 28,587,776 & 25,356 & 285,856 & 11,048 & 43.57 & 43.57\% \\
\bottomrule
\end{tabular}
\end{table*}

\subsection{Selector comparison}
\begin{table*}[t]
\centering
\scriptsize
\caption{Amplification by selector. Random baselines report mean $\pm$ sample standard deviation over 20 repetitions; the remaining selectors use seed 0.}
\label{tab:app-selectors}
\begin{tabular}{lrrrr}
\toprule
Selector & MV--WRN & VisA--WRN & MV--DINO & VisA--DINO \\
\midrule
Patch random & $0.99\pm0.09$ & $0.99\pm0.12$ & $0.99\pm0.04$ & $1.00\pm0.07$ \\
Image-balanced random & $0.99\pm0.09$ & $1.01\pm0.13$ & $0.99\pm0.04$ & $1.02\pm0.07$ \\
SimHash & 1.67 & 1.73 & 0.92 & 0.98 \\
Density top-$K$ & 0.00 & 0.00 & 0.11 & 0.01 \\
Sampled $k$-means medoid & 1.34 & 1.44 & 1.08 & 1.05 \\
Voronoi landmark mass & 0.41 & 0.33 & 0.30 & 0.29 \\
Local FF & 2.73 & 2.55 & 2.60 & 3.42 \\
Global FF & 16.41 & 31.17 & 23.83 & 43.57 \\
\bottomrule
\end{tabular}
\end{table*}
Global FF yields $A_K>1$ for every category in all four settings. The corresponding counts are 13/15, 11/12, 15/15, and 12/12 for Local FF; 12/15, 9/12, 9/15, and 6/12 for sampled $k$-means medoid; and 12/15, 11/12, 5/15, and 6/12 for SimHash.

\subsection{Achieved-radius certificate}
Let $S_K$ denote the returned centers and
\begin{equation}
  r_K=\max_{x\in\mathcal X}d(x,S_K).
\end{equation}
Define the relative contaminated cover number
\begin{equation}
  \mathcal C_r(\mathcal Q\mid\mathcal N)
  =\min_{C\subseteq\mathcal Q}
  \left\{|C|:\mathcal Q\subseteq B(\mathcal N,r)\cup B(C,r)\right\}.
\end{equation}
Since $S_K\cap\mathcal Q$ is feasible at the achieved radius,
\begin{equation}
  |S_K\cap\mathcal Q|
  \geq
  \mathcal C_{r_K}(\mathcal Q\mid\mathcal N).
\end{equation}
For
\begin{equation}
  U_K=\{q\in\mathcal Q:d(q,\mathcal N)>r_K\},
\end{equation}
a greedy packing $P_K\subseteq U_K$ with pairwise distances larger than $2r_K$ satisfies
\begin{equation}
  |P_K|
  \leq
  \mathcal C_{r_K}(\mathcal Q\mid\mathcal N)
  \leq
  |S_K\cap\mathcal Q|.
\end{equation}

\begin{table}[t]
\centering
\scriptsize
\caption{Fold-0 achieved-radius packing certificate for the same seed-0 Global FF memories as Table~\ref{tab:app-counts}.}
\label{tab:app-packing}
\begin{tabular}{lrrrr}
\toprule
Setting & $|U_K|/q$ & $|P_K|$ & $m_K$ & $|P_K|/m_K$ \\
\midrule
MVTec--WRN50 & 32.02\% & 95 & 2,065 & 4.60\% \\
VisA--WRN50 & 24.98\% & 142 & 2,247 & 6.32\% \\
MVTec--DINOv2 & 20.36\% & 70 & 12,160 & 0.58\% \\
VisA--DINOv2 & 26.92\% & 92 & 11,048 & 0.83\% \\
\bottomrule
\end{tabular}
\end{table}
The certificate is positive in all four settings. The proportion of selected contaminated centers directly certified at the achieved radius is $0.58\%$--$6.32\%$.

\section{Additional Results for Memory Interventions}
\label{app:downstream}

\subsection{Distribution of matched intervention gains}
The three-fold matched intervention improves P-AP in all 12 dataset--representation--fold macro comparisons and in 89 of 162 category--fold comparisons. The VisA--WRN50 win count is 13/36.

\begin{figure}[t]
  \centering
  \includegraphics[width=0.82\columnwidth]{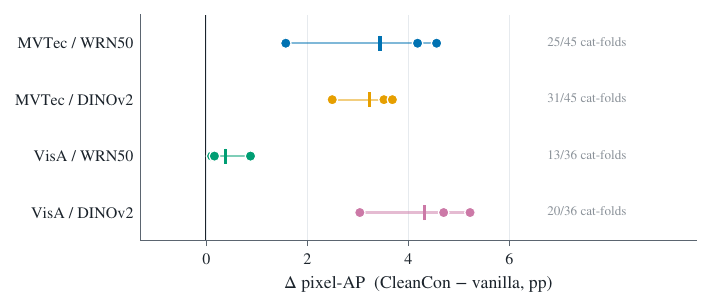}
  \caption{\textbf{Distribution of matched P-AP changes.} Points denote contamination folds, ticks denote three-fold means, and labels report category--fold wins.}
  \label{fig:app-matched-results}
\end{figure}

\subsection{Category-level retention results}
\begin{center}
\begin{tabular}{lcc}
\toprule
Setting & Categories with R80 $>$ R50 & Median $\Delta$P-AP \\
\midrule
MVTec--WRN50 & 9/15 & -- \\
MVTec--DINOv2 & 11/15 & $+0.407$ \\
VisA--WRN50 & 7/12 & $+0.145$ \\
VisA--DINOv2 & 11/12 & $+2.060$ \\
\bottomrule
\end{tabular}
\end{center}
Removing the largest positive category outlier on MVTec--WRN50 leaves a mean gain of $+0.292$. On VisA--WRN50, removing pcb4 changes the mean difference to $-0.156$. At fold level, R80 exceeds R50 in all 12 runs, and contamination is also higher at R80 in 11 of them.

\subsection{Metric consistency of the retention result}
\begin{table}[t]
\centering
\scriptsize
\caption{Change from R50 to R80. Values are differences of three-fold category-macro means; parentheses give the number of improving folds.}
\label{tab:app-retention-metrics}
\begin{tabular}{lrrrr}
\toprule
Setting & I-AUROC & P-AUROC & P-AP & AUPRO \\
\midrule
MVTec--WRN50 & $+1.239$ (3/3) & $+0.126$ (3/3) & $+0.929$ (3/3) & $+0.425$ (3/3) \\
MVTec--DINOv2 & $+0.242$ (3/3) & $+0.046$ (3/3) & $+0.521$ (3/3) & $+0.139$ (3/3) \\
VisA--WRN50 & $+1.071$ (3/3) & $+0.037$ (2/3) & $+0.875$ (3/3) & $+0.301$ (3/3) \\
VisA--DINOv2 & $+0.104$ (2/3) & $+0.028$ (2/3) & $+2.293$ (3/3) & $+0.120$ (2/3) \\
\bottomrule
\end{tabular}
\end{table}
The four-setting mean increases for all metrics. Fold agreement is strongest for P-AP and AUPRO; P-AUROC increases in two of three VisA folds.

\subsection{Clean filtering and normal-source expansion}
\label{app:source-expansion}
The six-category auxiliary subset contains cable, capsule, metal nut, pill, toothbrush, and zipper. On clean training data, applying the same 50\% image-retention rule reduces P-AP by 0.939 points. In the 15-category MVTec Addition-$0.10$ No-Overlap experiment, the normal-source expansion control increases $R_{\mathrm{src}}$ by 20.88 percentage points and changes P-AP by $-0.137\pm0.024$ points.

\begin{figure}[t]
  \centering
  \includegraphics[width=0.82\columnwidth]{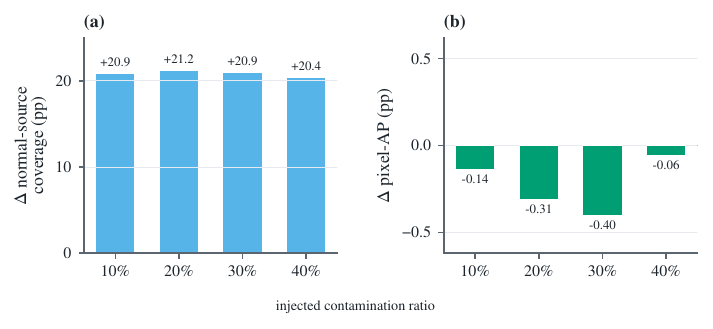}
  \caption{\textbf{Normal-source expansion.} The expansion control increases normal-source representation by approximately 21 percentage points without improving P-AP.}
  \label{fig:app-salvage}
\end{figure}

\subsection{Alternative candidate interventions}
We compare the whole-image median gate with four candidate controls. OOB-NN and projection-residual patch ranking retain patches from otherwise rejected images using their corresponding OOB scores. Oracle normal-region selection removes mask-positive patches. Normal-source expansion raises $R_{\mathrm{src}}$ at the same final memory size, while adaptive image retention replaces the fixed 50\% setting with a data-dependent retained fraction.
\begin{table*}[t]
\centering
\scriptsize
\caption{Alternative candidate interventions.}
\label{tab:app-alternatives}
\begin{tabularx}{\textwidth}{l l X X}
\toprule
Candidate rule & Selection unit & Experimental scope & Result \\
\midrule
OOB-NN patch ranking & patch & six-category auxiliary experiment & lower local ranking and P-AP than projection-residual patch ranking \\
Projection-residual patch ranking & patch & six-category auxiliary experiment & $+1.233$ P-AP over OOB-NN, $-0.371$ relative to whole-image gating \\
Oracle normal-region selection & patch & ground-truth mask setting & removes anomalous regions and retains only normal regions as candidates \\
Normal-source expansion & image + patch & 15 MVTec categories, three No-Overlap folds & $-0.137$ P-AP and $-0.215$ AUPRO relative to whole-image gating \\
Adaptive image retention & image & Exact-Final, seed 0 & $+0.601$ P-AP at $5\%$ and $-2.095$ at $10\%$ relative to whole-image gating \\
\bottomrule
\end{tabularx}
\end{table*}
Projection-residual patch ranking exceeds OOB-NN patch ranking but remains below the whole-image median gate. Lowering the patch quantile from Q0.95 to Q0.90 reduces contamination while changing P-AP from 67.921 to 67.874.

\subsection{Risk-aware coverage control}
In a fold-0, 15-category MVTec setting, we modify the coverage builder to use the OOB risk score. The traversal oversamples coverage candidates and replaces a subset of high-risk centers.
\begin{table}[t]
\centering
\scriptsize
\caption{Risk-aware coverage control.}
\label{tab:app-risk-kcenter}
\begin{tabular}{llrrrr}
\toprule
Mixed-train contamination & Builder & $p_{\mathrm{mem}}$ & I-AUROC & P-AP & AUPRO \\
\midrule
$5\%$ & Global FF & $0.0155\%$ & 99.641 & 72.848 & 95.976 \\
       & risk-aware & $0.0093\%$ & 99.596 & 72.814 & 95.974 \\
$10\%$ & Global FF & $0.8059\%$ & 99.507 & 70.415 & 94.323 \\
        & risk-aware & $0.6934\%$ & 99.519 & 70.573 & 94.389 \\
\bottomrule
\end{tabular}
\end{table}
The risk-aware builder reduces $p_{\mathrm{mem}}$ at both ratios. P-AP changes from 72.848 to 72.814 at $5\%$ and from 70.415 to 70.573 at $10\%$.

\subsection{Detector scaffolds in the SoftPatch-style control}
The SoftPatch-style comparison uses all 15 MVTec and 12 VisA categories at contamination fold 0 and seed 0. Within WRN50 rows, both arms share PatchCore inference; within DINOv2 rows, both share ProCon inference. The SoftPatch-style arm uses single-discriminator selection and reliability weighting, while the \method arm uses OOB image gating.

\begin{table*}[t]
\centering
\scriptsize
\caption{SoftPatch-style control and \method under matched detector scaffolds.}
\label{tab:app-softpatch}
\setlength{\tabcolsep}{3.1pt}
\begin{tabular}{llrrrrrrr}
\toprule
Dataset & Backbone / ratio & \multicolumn{3}{c}{SP-style} & \multicolumn{3}{c}{\method} & $\Delta$P-AP \\
\cmidrule(lr){3-5}\cmidrule(lr){6-8}
& & I-AUROC & P-AP & AUPRO & I-AUROC & P-AP & AUPRO & \\
\midrule
MVTec & WRN50 / 5\%   & 98.241 & 60.029 & 93.465 & 97.539 & 61.016 & 93.539 & +0.987 \\
MVTec & WRN50 / 10\%  & 98.025 & 60.013 & 92.971 & 97.762 & 60.691 & 93.827 & +0.679 \\
MVTec & DINOv2 / 5\%  & 98.880 & 70.570 & 95.440 & 99.147 & 72.255 & 95.770 & +1.686 \\
MVTec & DINOv2 / 10\% & 98.856 & 68.890 & 94.814 & 99.316 & 72.052 & 96.063 & +3.162 \\
VisA & WRN50 / 5\%    & 92.810 & 41.495 & 93.735 & 92.596 & 40.061 & 93.693 & -1.434 \\
VisA & DINOv2 / 5\%   & 97.502 & 46.147 & 96.253 & 97.712 & 49.522 & 96.861 & +3.374 \\
\bottomrule
\end{tabular}
\end{table*}

On VisA--WRN50, pooled final-memory contamination is approximately $0.108\%$ for the SP-style control and $0.006\%$ for \method, while P-AP is 41.495 and 40.061, respectively.

\end{document}